\documentclass[letterpaper, 10 pt, conference]{ieeeconf}  %

\IEEEoverridecommandlockouts                              %

\newcommand{\etal}{\emph{et al.}}
\usepackage[noadjust]{cite}

\usepackage{multicol}
\usepackage[bookmarks=true]{hyperref}
\usepackage{tikz}
\usepackage{graphicx}

\usepackage{textcomp}
\usepackage{color}
\usepackage{mathtools}
\usepackage{amsmath}
\usepackage{bm}

\usepackage{booktabs}  %
\usepackage{multirow}  %

\usepackage{epsfig,graphicx,amsmath,amsfonts}
\usepackage{transparent}
\usepackage{upgreek}

\newcommand{\E}{\mathbb{E}}  %
\newcommand{\R}{\mathbb{R}}  %
\newcommand{\T}{\top}  %
\def\ith{i^\textit{th}}
\def\bmatrix#1{\left[ \begin{matrix} #1 \end{matrix} \right]}  %
\newcommand{\bfC}{\ensuremath{\bm{C}}}
\newcommand{\bfI}{\ensuremath{\bm{I}}}
\newcommand{\bfJ}{{\ensuremath{\bm{J}}}}

\newcommand{\bfM}{{\ensuremath{\bm{M}}}}
\newcommand{\bfa}{\ensuremath{\bm{a}}}

\newcommand{\bfd}{\ensuremath{\bm{d}}}
\newcommand{\bfg}{\ensuremath{\bm{g}}}
\newcommand{\bfq}{\ensuremath{\bm{q}}}
\newcommand{\bff}{\ensuremath{\bm{f}}}
\newcommand{\bfell}{\ensuremath{\bm{\ell}}}
\newcommand{\bfp}{\ensuremath{\bm{p}}}
\newcommand{\bfs}{\ensuremath{\bm{s}}}
\newcommand{\bfx}{\ensuremath{\bm{x}}}
\newcommand{\bftau}{\ensuremath{\bm{\tau}}}
\newcommand{\calL}{{\ensuremath{\cal{L}}}}
\newcommand{\calP}{{\ensuremath{\cal{P}}}}
\newcommand{\calR}{{\ensuremath{\cal{R}}}}
\newcommand{\eqnref}[1]{Eq.~\eqref{#1}}
\newcommand{\eqnsref}[2]{Eqs.~\eqref{#1} and~\eqref{#2}}

\newcommand\numberthis[1]{\addtocounter{equation}{1}\tag{\theequation}\label{#1}}
\usepackage[normalem]{ulem}
\usepackage{multirow}
\usepackage{float}
\usepackage[utf8]{inputenc}
\usepackage[TS1,T1]{fontenc}
\usepackage{array, booktabs}
\usepackage{xcolor}

\title{\LARGE \bf
\huge \bf Joint Space Control via Deep Reinforcement Learning
}

\author{Visak Kumar,$^{\dagger}$\thanks{$^\dagger$Also affiliated with Georgia Tech.  Work was performed while the first author was an intern with NVIDIA.  $^*$The initial version of the approach was designed and implemented by the second author.} David Hoeller,$^*$ Balakumar Sundaralingam, Jonathan Tremblay, Stan Birchfield \\
NVIDIA
}

\begin{document}

\maketitle
\thispagestyle{empty}
\pagestyle{empty}

\begin{abstract}
The dominant way to control a robot manipulator uses hand-crafted differential equations leveraging some form of inverse kinematics / dynamics.
We propose a simple, versatile joint-level controller that dispenses with differential equations entirely.
A deep neural network, trained via model-free reinforcement learning, is used to map from task space to joint space.
Experiments show the method capable of achieving similar error to traditional methods, while greatly simplifying the process by automatically handling redundancy, joint limits, and acceleration / deceleration profiles.
The basic technique is extended to avoid obstacles by augmenting the input to the network with information about the nearest obstacles.
Results are shown both in simulation and on a real robot via sim-to-real transfer of the learned policy.
We show that it is possible to achieve sub-centimeter accuracy, both in simulation and the real world, 
with a moderate amount of training.\let\thefootnote\relax\footnotetext{Video is at \url{https://youtu.be/ICfve-GTTp8}}

\end{abstract}

\section{Introduction}

Fundamental to robot control is the mapping from \emph{task space} to \emph{joint space}.
With this mapping, a desired task described in the Cartesian world in which the robot moves can be translated into specific low-level commands to be sent to the joint motors to accomplish the task.
The control law that maps Cartesian task space target poses to robot joint space commands is known as \emph{operational space control}~\cite{khatib1987ra:osc}. 

For decades, traditional approaches have leveraged analytic techniques based on differential equations to perform operational space control for reaching end-effector target poses~\cite{schaalosc}.
The prebuilt controllers of commercially available robotic systems fall into this category. 
Such approaches, however, are unable to navigate in unstructured environments, because they cannot reason about constraints in joint space (\emph{e.g.}, avoiding collision between the robot's links and the environment). 
This limitation greatly reduces the usefulness of pre-built operational space controllers. 
Recent methods~\cite{rmp,cheng2018rmpflow,kappler_reactive_2018} that integrate perception with control have shown success in reaching task space goals while avoiding dynamic obstacles. Such methods, however, require extensive robotics expertise to integrate into new manipulators, and they also require tuning for novel environments or tasks.

Reinforcement learning (RL) is a promising approach to replace analytic, hand-crafted solutions with learned policies~\cite{duan2016icml:bench}. 
Most RL-based robotic systems learn in task space rather than in joint space~\cite{Shao-RSS-20,migimatsu_bohg_2020,martin2019variable,lee2019icra:vistouch}. 
By working in task space (\emph{i.e.}, Cartesian end-effector space), the learning problem is greatly simplified, and the safe operation of the robot can be delegated to an analytic low-level controller that maps the RL actions to joint commands. 
Like the pre-built controllers mentioned above, however, such solutions cannot operate in unstructured environments in which collisions between the arm and environment require joint-level reasoning.

\begin{figure}
    \centering
    \includegraphics[width=0.6\linewidth, trim= 0 1cm 0 0,clip]{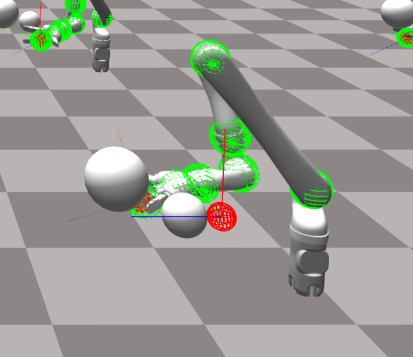} 
    \caption{
    Our deep RL-based JAiLeR system
    is able to learn a policy that enables a robot manipulator to reach any goal position (red sphere) in the workspace with sub-centimeter average error, while avoiding obstacles (silver spheres).  The green points on the robot links are used for distance computations for collision avoidance.}
    \label{fig:jaco_obs}
\end{figure}

Several research efforts have used reinforcement learning to control a robot manipulator at the joint level, both in simulation~\cite{brockman2016openai,schulman2017proximal,pong2018corl:tempdiff} and in the real world~\cite{levine2015,sadeghi2018cvpr:viv,zhu2018rss:rlil,mahmood2018corl:bench,luo2020ijcnn:accrl}.
These have involved learning to reaching a point in space and, in some cases, incorporating perception to accomplish a task.
While much progress has been made, motion of the robot joints is often not smooth, and the early termination of the learning procedure does not guarantee that the robot will cease to move once the goal is reached.
Furthermore, results are often reported in terms of reward function values or task success, and tasks often restrict the robot motion to a small portion of the workspace, thus making it difficult to assess geometric accuracy in a larger context.

\begin{figure*}
    \centering
    \includegraphics[width=0.7\linewidth]{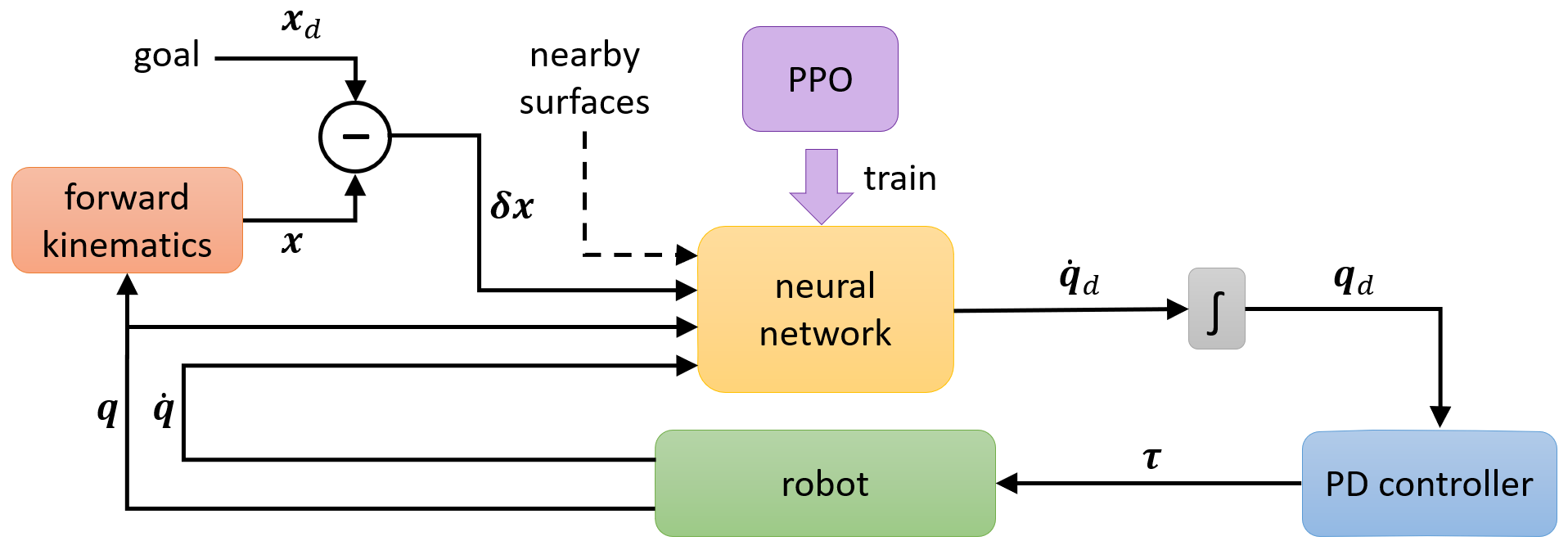} 
    \caption{Overview of the proposed JAiLeR approach.  The goal is provided externally (at run time) or by curriculum learning (during training).  The robot can be either simulated or real.  To avoid clutter, the inputs ($\delta \bfx$, $\bfq$, and ${\dot{\bfq}}$) to PPO used to train the network are not shown.  Information about nearby surfaces may optionally be fed as input to the neural network, for obstacle avoidance.}
    \label{fig:overview}
\end{figure*} 

It remains to be seen, then, whether deep reinforcement learning can be used to accurately, smoothly, and stably (with stopping) control a high-DoF robot to reach within a large workspace.
If this question could be answered in the affirmative, then learned controllers could begin to replace traditional controllers for large-scale, real-world tasks.
An advantage of such learned controllers over traditional controllers is that they can more easily be connected with deep neural network-based perception approaches, since they are networks themselves.

In this paper, we take a step in this direction by presenting Joint Action-space Learned Reacher (JAiLeR), a method that leverages recent advancements in model-free deep reinforcement learning to directly map task-space goals into joint-space commands.  See Fig.~\ref{fig:jaco_obs}.
Redundancy, joint limits, and acceleration / deceleration profiles are automatically handled by learning directly from behavior in simulation, without the need to explicitly model the null space or manually specify constraints or profiles.
Collision avoidance is incorporated into the approach by augmenting the network's input with distances between robot links and nearby surfaces, which allows the robot to avoid dynamic obstacles while still reaching target end-effector poses---without requiring scene-specific training.
The proposed approach thus provides a much simpler alternative to classical approaches to obstacle avoidance, such as control barrier functions~\cite{ames2019ecc:cbf} or Riemannian motion policies~\cite{rmp}.
We show that JAiLeR can reach a large workspace with average error less than 1~cm, both in simulation and in reality.

\section{Learning to control}

The goal of this work is to investigate the possibility of replacing operational space controllers (described in the Appendix for reference) with a learned controller.
Given a large amount of training data, a neural network with sufficient capacity, and a reasonably efficient training algorithm, we will show that a controller can be learned to perform the mapping between task and joint spaces, with comparable accuracy to hand-crafted controllers.
A learned controller provides several advantages over traditional approaches, namely, that redundancy, joint limits, and acceleration / deceleration profiles are automatically learned from behavior in simulation during training.

In our approach, a neural network executes a policy 
$\pi_\theta~:~\bfs_t~\mapsto~\bfa_t$ 
that maps the current state $\bfs_t$ to an action $\bfa_t$, where $\theta$ are the parameters of the policy (\emph{i.e.,} the network weights).  
During training, the network learns to maximize the expected total reward, $\arg \max_\theta \E \left[ \sum_{t=1}^T r(\bfs_t,\bfa_t)\right]$, where $r:\bfs_t,\bfa_t \mapsto \R$ is the reward function, and $T$ is the episode length.

An overview of our system is illustrated in Fig.~\ref{fig:overview}.
The end effector error $\delta \bfx = \bfx_d - \bfx$ in task space is fed to the network, along with the robot's current joint angles $\bfq$ and velocities ${\dot \bfq}$, as well as (optionally) information about nearby surfaces.
The network outputs the desired joint velocities ${\dot \bfq}_d$, which are then integrated to obtain the desired angles $\bfq_d$.
These are fed to a proportional-derivative (PD) controller, which outputs the torques to control the robot.

\subsection{State and action representation}

The choice of state and action space is crucial.  One possibility is to represent both states and actions in task space, as done in \cite{Shao-RSS-20,migimatsu_bohg_2020,martin2019variable,lee2019icra:vistouch}.  In this case the policy outputs~$\bfx_d$, ${\dot \bfx_d}$, or ${\ddot \bfx_d}$, which is then leveraged by an analytic controller\footnote{Such as \eqnref{eq:osc_vel_law} or \eqnref{eq:osc_acc_law} from the Appendix.}  to compute the joint torques. The problem with this approach is that the policy cannot directly account for task constraints in manifolds other than the end-effector's Cartesian task space. In particular, it cannot handle collision avoidance of links other than the end effector.  This is a major limitation for deploying such reinforcement learning policies in unstructured environments.

In contrast, we propose to learn the mapping from task space to joint space, as shown in the figure.  
That is, the network maps $\delta \bfx$ to ${\dot \bfq}_d$, with the current joint angles and velocities providing context to the network.
We choose to output desired velocities, since the integration to desired positions provides robustness to noise, and since position-based control yields better transferability from simulation to reality since it abstracts away the dynamics.
The desired joint targets are fed to a  proportional-derivative (PD) controller: \begin{align*}
\bftau &=  \framebox{\ensuremath{\bfC {\dot \bfq} + \bfg}} + k_p (\bfq_d - \bfq) + k_d (\dot{\bfq}_d - \dot{\bfq}), \numberthis{eq:osc_law}
\end{align*}
where the robot dynamics (contained in the terms $\bfC {\dot \bfq} + \bfg$) are a black box inaccessible to the user.
In other words, the robot's internal controller computes the torques from the input $k_p(\cdot) + k_d(\cdot)$, where $\bfq_d=\int {\dot \bfq}_d$.  Note that there is no $\bfM\ddot{\bfq}_d$ term, because setting desired acceleration will lead to poor performance if there are modeling errors in $\bfM$.

Alternatively, we could output desired positions, accelerations, or torques.
There are pros and cons to each of these approaches, and we have tried each of them with varying degrees of success.
We have settled on position-based controllers because, although they yield slightly less smooth behavior than the others, they are much more easily transferred to the real robot.
Moreover, they do not require direct access to the motor torques, in contrast to analytic controllers, which can only be used on robotic platforms that provide such direct access.

To facilitate obstacle avoidance, the input can be augmented with information about the closest objects.  
Specifically, let $\calL_i$ be the (possibly infinite) set of points on the surface of the $\ith$ link, and let $\calP$ be the (possibly infinite) set of points on the surface of all objects in the world.
Let $\bfp_i \in \calP$ and $\bfell_i \in \calL_i$ be the (possibly non-unique) closest points in the two sets, \emph{i.e.},
\begin{align}
\| \bfp_i - \bfell_i \| = \min_{\bfp \in \calP} \min_{\bfell \in \calL_i} \| \bfp - \bfell \|. \numberthis{eq:closestpoints}
\end{align}
The network receives as input, in addition to the quantities already mentioned, the vector for each link connecting these closest points,  \emph{i.e.}, $\bfd_1, \bfd_2, \ldots, \bfd_n$, where $\bfd_i \equiv \bfp_i - \bfell_i$, and $n$ is the number of links.
In practice, we simplify the above by considering, for each link $i$, only the points $\calL_i$ on a sphere attached to the link, as shown in Fig.~\ref{fig:jaco_obs}.
Note that this formulation is independent of the spatial resolution of the robot links, and with proper scene representation it scales sublinearly with the spatial resolution of the scene.

This approach to capturing information about obstacles is advantageous because the network does not need to be retrained when the environment changes.
In addition, the network is independent of the scene parameterization.  
Alternatively, of course, we could extend the neural network with additional layers (such as PointNet~\cite{qi2017cvpr:pointnet,qi2017nips:pointnetpp}) to automatically process the scene rather than relying on a separate traditional geometric computation as described in \eqnref{eq:closestpoints}.
A depth camera, coupled with the robot's forward kinematics, can be used for such purposes, although perception is outside the scope of this paper.

To summarize, the network receives an $(m+2n)$-dimensional input vector $\bfs=\bmatrix{\delta \bfx^\T, \bfq^\T, {\dot \bfq}^\T}^\T$ without obstacle avoidance, or an $(m+5n)$-dimensional input vector $\bfs=\bmatrix{\delta \bfx^\T, \bfq^\T, {\dot \bfq}^\T, \bfd_1^\T, \ldots, \bfd_n^\T}^\T$
with obstacle avoidance. 
The value $m$ is determined by the state representation:  $m=3$ in the case of position-only, \emph{i.e.}, $\bfx \in \R^3$; $m=6$ or~7 in the case of position and orientation, \emph{i.e.}, $\bfx \in \mathbb{SE}(3)$, depending upon whether quaternions are used; and so forth.
Either way, the action is an $n$-dimensional output vector
$\bfa={\dot \bfq_d}$.

\subsection{Reward}
\label{sec:reward}

The reward used to train the network is straightforward:
\begin{align}
r(\bfs,\bfa) = \exp(-\lambda_{\textit{err}} \| \delta \bfx \|^2) - \lambda_{\textit{eff}} \| {\ddot \bfq} \| - \lambda_{\textit{obs}} \sum_{i=1}^n \psi_i,
\numberthis{eq:reward}
\end{align}
where the first term encourages the end effector error $\delta \bfx$ to go to zero, the second term encourages solutions that require minimal effort, and the third term penalizes the robot for getting too close to an obstacle.
Note that minimizing effort causes the network to automatically handle joint redundancies by moving the joints as little as necessary to achieve the task.
The obstacle avoidance penalties are given by
\begin{align}
\psi_i = \max\left(0, 1 - \| \bfd_i \| / d_{\textit{max}}\right),
\end{align}
where $d_{\textit{max}} = 5$~cm is the maximum distance that incurs a penalty.
We set the relative weights according to a linear sweep as follows: 
$\lambda_{\textit{err}}=20, \lambda_{\textit{eff}}=0.005$, and 
$\lambda_{\textit{obs}}=0.1$.

When the network is trying to reach a position in 3D space, $\delta \bfx$ is simply the Euclidean distance between the current position and the goal position.
When the network is trying to reach a specific position and orientation, $\delta \bfx$ in \eqnref{eq:reward} is the sum of three Euclidean distances, using three points defined along orthogonal axes from the goal, as described in~\cite{molchanov2019iros:s2mr,zhou2019cvpr:continrot}.

\subsection{Learning procedure}
\label{sec:learn-arch}

We train the network using PPO~\cite{schulman2017proximal}, a recent model-free deep reinforcement learning algorithm.
In our system, the network has 3~layers, each with 128~neurons with \texttt{tanh} activation for the first two layers.  
The network is trained using curriculum learning~\cite{oudeyer2007tec:intrinsic,bengio2009icml:currlearn,krueger2009c:flexshape,kumar2010nips:selfpaced,lee2011cvpr:easylearn,florensa2017corl:reversecurr,hacohen2019icml:power,matiisen2020tnnls:teacherstudent,2018-TOG-deepMimic,portelas2020ijcai:currlearn,narvekar2020jmlr:currlearn,luo2020ijcnn:accrl}, where the entire workspace is divided into a sequence of regions $\calR_1 \subset \calR_2 \subset \cdots \subset \calR_k$, where $\calR_k$ is the workspace, and $k$ is the number of regions in the curriculum.
Starting with $i=1$, the network is trained on region $\calR_i$, periodically measuring the average error ${err}_{avg}$, until ${err}_{avg} < {th}$, at which point training proceeds to region $\calR_{i+1}$, and so forth; where ${th}=1$~cm is a threshold.
Note that each region completely contains the previous region, to prevent catastrophic forgetting.
This does not lead to sample complexity problems, since the region sizes scale slowly (non-exponentially), and simulation provides us with ample samples.
We found curriculum learning to be crucial for getting good results.

\section{Experiments}

In this section we evaluate our JAiLeR system in both simulation and in the real world.  

\subsection{Reaching policy}

We used the procedure described above to train a reaching policy in simulation using a Kinova Jaco 6-DoF robot arm.
Although it is straightforward to incorporate orientation constraints into the method, all experiments in this paper were conducted without regard for orientation. 
Using the simulation system developed by Liang et al.~\cite{liang2018corl:gpudrl}, we trained 40 robots in parallel, which greatly sped up the training time.  On an NVIDIA Titan Xp machine, the policy required about two hours to train from scratch.

\begin{figure}
    \centering
    \begin{tabular}{cc}
    \hspace{-0.9em}    \includegraphics[width=0.48\columnwidth]{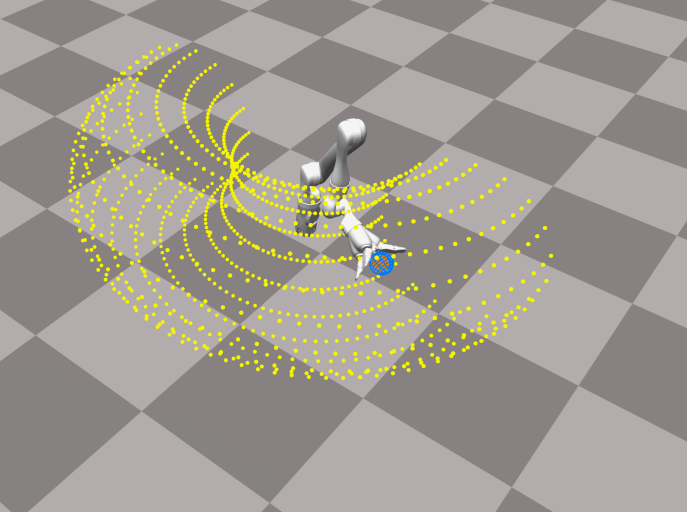} & 
    \hspace{-1.2em} \includegraphics[width=0.48\columnwidth]{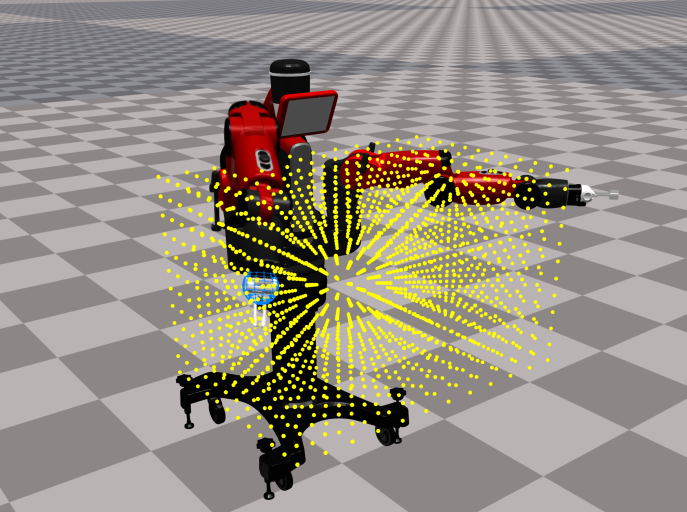} \\
    (a) Jaco workspace & (b) Baxter workspace
    \end{tabular}
    \caption{Workspace for Jaco and Baxter reaching experiments.  Due to Jaco's rotational symmetry, the policy can reach the entire $360^\circ$ space (not shown), with a simple offset trick at runtime.}
    \label{fig:jaco_bax_workspace}
\end{figure}

The workspace is defined as a torus with major radius 45~cm and minor radius 30~cm, centered at the base of the robot, as shown in Fig.~\ref{fig:jaco_bax_workspace}a.  
For curriculum learning, the workspace is divided into 4 regions, each a superset of the previous.  Each region $\calR_i$ is a partial torus, that is, a surface of revolution in which the rotation angle is less than 360~degrees.  All four regions share their major radius.  The minor radii are given by 15, 20, 30, and 30~cm, respectively; and the rotation angles are given by $90^\circ$, $120^\circ$, $120^\circ$, and $180^\circ$. 
Only the final $\calR_4$ is shown in the figure.

During training, the goal end-effector locations are sampled uniformly within the current curriculum region. The start location is the position reached at the end of the previous rollout, without resetting the robot. We terminate the rollout and reset the robot to a pre-defined home position if the distance between the robot end-effector and goal increases greater than $80$ cm (this termination condition is only needed in curriculum region $\calR_1$). 
Each rollout is 8 seconds long, with a time step of 0.01~sec. 
For the 40 robots simulated in parallel, we use the same policy to sample actions for all robots. Once the data from each robot is collected, the policy is updated by sampling batches from the entire dataset, thereby using data from each robot to improve the policy.

The trained policy was tested in simulation by randomly sampling start and goal positions within the $\calR_4$ workspace (using a uniform distribution).  
Our JAiLeR system successfully reached within 1.0~cm of 96.5\% of the goal positions, with 0.4~cm average error.  
For all of these successful runs, the mean time to completion was 14.3 seconds, with a standard deviation of 2.2 seconds.
Tab.~\ref{tab:jaco_reach_sim} shows these results compared with two operational space controllers (described in the Appendix).  JAiLeR outperforms OSC-V, and it performs comparably to OSC-A.  Note that the curriculum learning strategy is crucial to success.

\setlength{\tabcolsep}{5pt}
\begin{table}[]
    \centering
    \caption{Jaco reaching experiment in simulation for Region $\calR_4$. \\ OSC-V and OSC-A are described in the Appendix.
    }
    \begin{tabular}{ccccc}
        \toprule
        & without \\
        & curriculum & JAiLeR & OSC-V & OSC-A \\
        \midrule
        success@1.0~cm & 6.6\% & 96.5\% & 53.1\% & 97.9\% \\
        average error (cm) & 2.9 & 0.4 & 0.8 & 0.4 \\
        completion time (s) & -- & 14.3$\pm$2.2 & 20.8$\pm$3.6 & 13.8$\pm$2.3 \\
    \bottomrule
\end{tabular}
    \label{tab:jaco_reach_sim}
\end{table}
\setlength{\tabcolsep}{6pt} %

Additionally, we observed that OSC-A is unable to stably reach poses that are near kinematic singularities.
For example, if the goal position is set near the edge of the reachable workspace, the robot will stretch out to align the links, then oscillate continually.  To overcome such a problem, practitioners often apply a large weight on the null space, but this requires delicate implementation and careful tuning. 
Our learned JAiLeR policy, on the other hand, stably handles poses near singularities, because the smooth activation functions of the neural network allow it to represent a smooth mapping.

Although the work of Lewis et al.~\cite{lewis2019tr:drl} is primarily an analysis of the generalization failures of naive deep reinforcement learning for control, it is nevertheless instructive to compare our results with theirs.
Whereas their procedure uses a fixed start configuration and samples only goals, we sample start-goal pairs from within the workspace, so that our system is able to reach from any start $\bfx_s \in \calR_k$ to any goal $\bfx_g \in \calR_k$ within the workspace.
Whereas their system terminates as soon as the end effector reaches within an $\epsilon$-ball of the goal, we require the end effector to settle down to zero velocity before measuring error.
And whereas we achieve an average error of 0.4~cm over a large workspace, their system exhibits errors greater than $\epsilon=10$~cm for more than half of the sampled goals, for moderately-sized workspaces, even when only half the joints are used.
Our purpose in this comparison is to highlight the significance of our results, namely, that geometrically accurate, robust, stable joint-level control is possible using an existing deep RL algorithm.

\begin{figure}
    \centering
    \includegraphics[width=0.97\columnwidth]{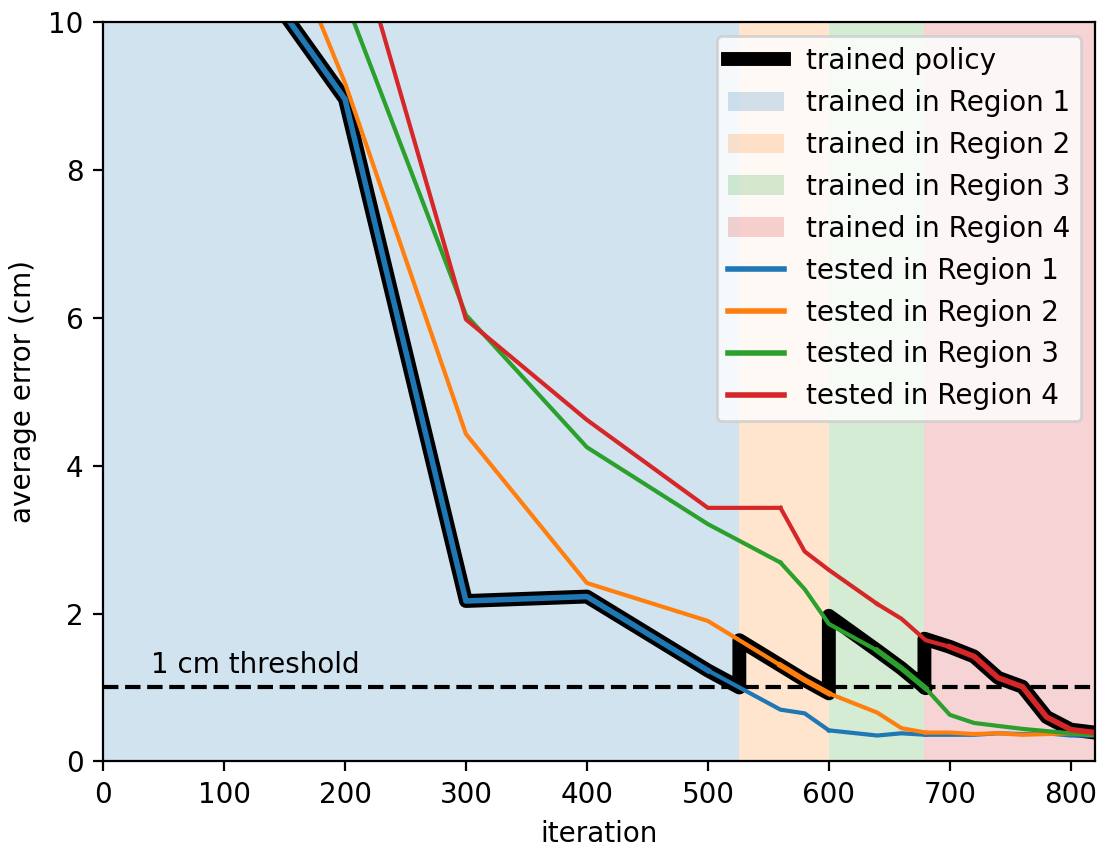} 
    \caption{Curriculum learning through four regions 
    $\calR_i \subset \calR_{i+1}$, $i=1,2,3$, where $\calR_i$ is Region $i$.  
    Transition from one region to the next occurs when the policy accuracy reaches the threshold of 1~cm.
    The network shows good generalization from smaller training sets to larger enclosing test sets.}
    \label{fig:currprog}
\end{figure}

Curriculum learning is shown in Fig.~\ref{fig:currprog}.  
Training the network involves progressing through a sequence of four curriculum stages with increasingly larger regions.  Switching to the next region occurs when the average error reaches a threshold of 1~cm.  
Note from the figure that the method shows surprisingly good generalization when the policy trained on a smaller region is tested on a larger enclosing region.  
Moreover, as the policy improves on the training region, it also improves on the larger enclosing regions.

\subsection{Obstacle avoidance}

The network was augmented with inputs regarding nearby obstacles, as described earlier.  
We used transfer learning to initialize the weights with values from the non-obstacle-aware network trained in the previous subsection.
During obstacle-aware training, 1--3 virtual spheres with radius 8~cm were placed randomly within the workspace so that they remained at least 5~cm from both the start and goal. 
Otherwise, training proceeded exactly as before, with the same curriculum.
The network learned to avoid obstacles, with a moderate impact on accuracy.
When tested with $n_{test}=1,2,3$ randomly placed spheres, the policy avoided collision, with performance degrading only slightly as the number of obstacles increases.
See Table~\ref{tab:obsavoid}, where the ``force ratio'' is the ratio of the normal to tangential force when the robot collides with an obstacle.  Thus, a small ratio indicates that the robot grazed the obstacle, rather than causing a hard collision.
The average endpoint error remained below 1~cm.

\begin{figure}
    \centering
    \begin{tabular}{cccc}
    \hspace{-0.9em}    \includegraphics[width=0.48\columnwidth]{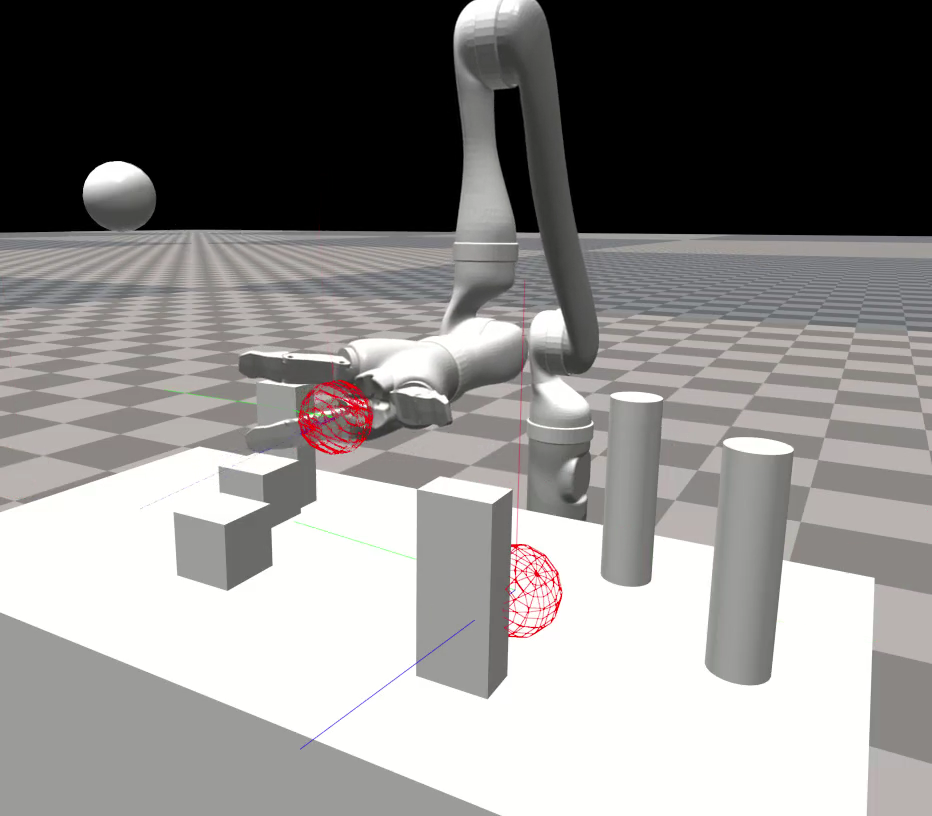} & 
    \hspace{-1.0em}    
    \includegraphics[width=0.48\columnwidth]{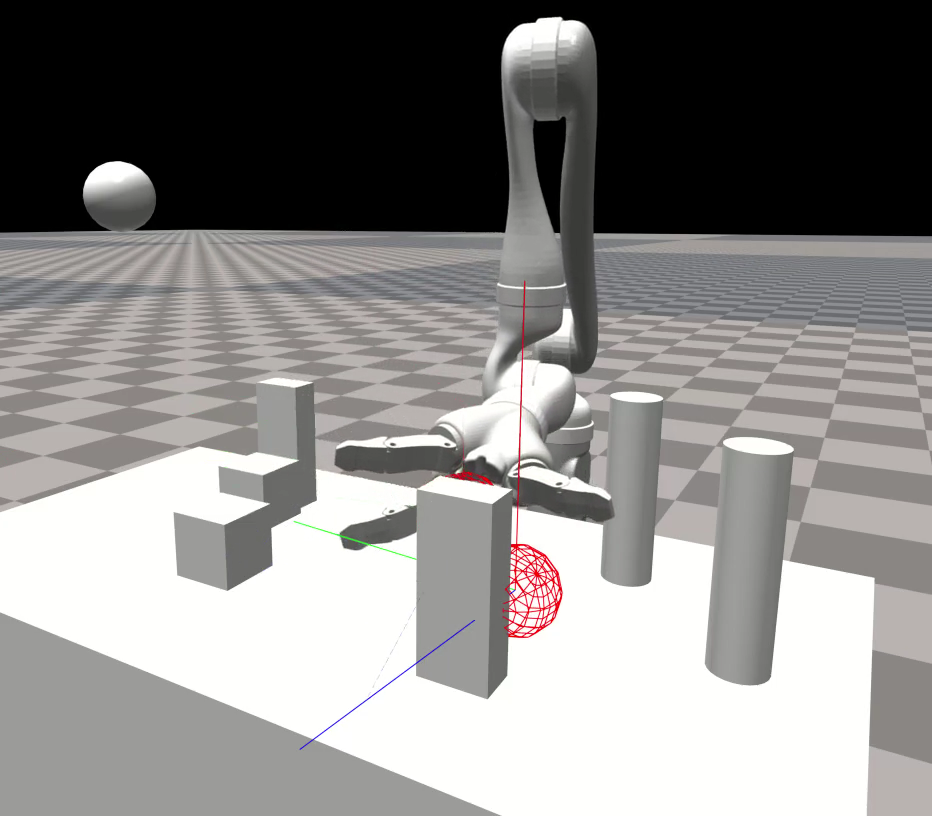} 
    \end{tabular}
    \caption{By feeding the network additional inputs (vectors pointing from links to the nearest obstacles), the policy learns to avoid obstacles.  This is \emph{without} scene-specific training, because the policy is trained only on spherical obstacles.
    The Jaco robot is reaching for a target (large red sphere).}
    \label{fig:jaco_obstacle_avoid}
\end{figure}

Due to our geometric-based formulation, the policy trained on spheres automatically generalizes to other shapes.
Fig.~\ref{fig:jaco_obstacle_avoid} shows the robot reaching for a target while avoiding obstacles.
These results show the potential of the proposed approach to incorporate sensory feedback at run time.

\begin{table}[]
    \centering
    \caption{Jaco obstacle avoidance experiment in simulation.  The force ratio shows mean $\pm$ std.}
    \begin{tabular}{cccc}
        \toprule
        no. obstacles & no. trials & collision-free & force ratio \\
        \midrule
 1 & 350 & 95.4\% & 0.27 $\pm$ 0.08 \\
 2 & 350 & 91.7\% & 0.41 $\pm$ 0.14 \\
 3 & 350 & 86.6\% & 0.63 $\pm$ 0.12 \\
    \bottomrule
\end{tabular}
    \label{tab:obsavoid}
\end{table}

\subsection{Sim-to-Real transfer}

To validate sim-to-real, we trained a reaching policy for the Baxter robot using the procedure described earlier.  
The workspace for the Baxter was defined as a $30 \times 40 \times 30$~cm region, shown in Fig.~\ref{fig:jaco_bax_workspace}b.
The curriculum followed a similar strategy as before.
Although the Baxter is capable of position- or velocity-based control, we observed that the latter is noisy and unstable, due to noise in the encoders and other effects.
As a result, we do not send velocities directly to the robot, rather relying upon position-based control.

The policy trained only in simulation works on the real robot.
The only change we had to make was to decrease the proportional gain in the controller to reduce oscillations, and to apply an exponential filter to smooth the encoder readings.
The reference position fed to \eqnref{eq:osc_law} was computed from the desired velocity as
$\bfq_r = \bfq + \lambda_1 \delta_t {\dot \bfq}_d$,
where $\lambda_1 = 0.5$,  %
and $\delta_t=0.01$~sec is the temporal sampling interval.

To test the policy on the real robot, we sampled goal positions sequentially (within the workspace described above) so that the robot never had to reset to a start position.
Rather, it simply moved sequentially through the goal positions, aiming for the next goal after the previous one was reached.
Over this workspace, the real Baxter achieved an average error of 0.9~cm, with a standard deviation of 0.6~cm.
Note, for comparison, that Rupert et al.~\cite{rupert2015humanoids:mpc,terry2017humanoids:mpc} achieved $1$ to $2.5$~cm steady-state error applying a traditional controller to the Baxter robot.
Our JAiLeR approach thus achieves results comparable to those of traditional controllers.

\begin{figure}
    \centering
    \includegraphics[width=0.97\columnwidth]{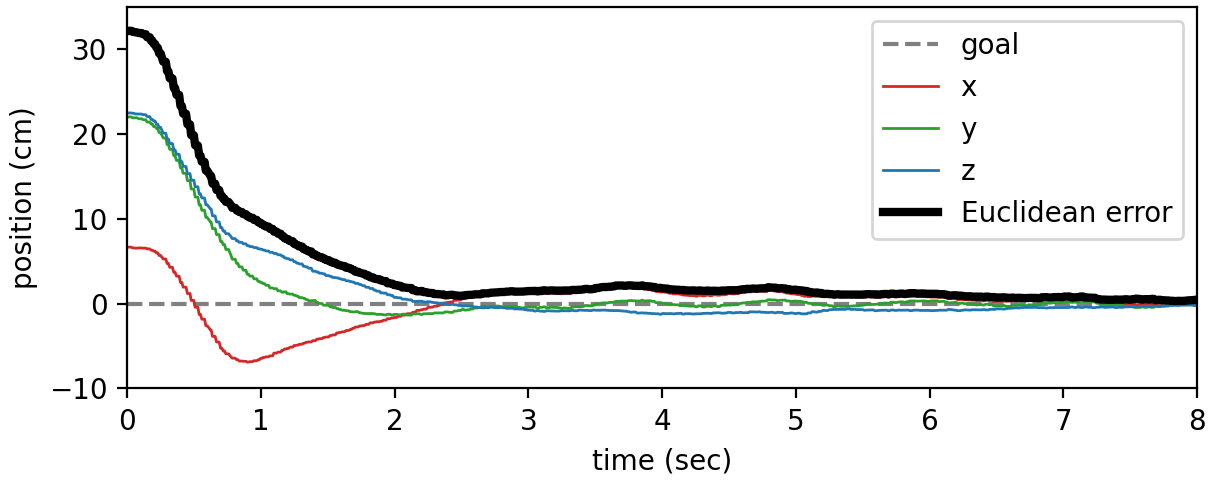} 
    \caption{Step response of the real Baxter robot reaching for a goal, showing the speed, accuracy, smoothness, relatively small overshoot, and stability of our learned policy. This policy was not fine-tuned in the real world and comes directly from simulation. }
    \label{fig:baxter_time_plot}
\end{figure}

\begin{figure}
    \centering
    \begin{tabular}{cc}
    \hspace{-0.9em}    \includegraphics[width=0.48\columnwidth]{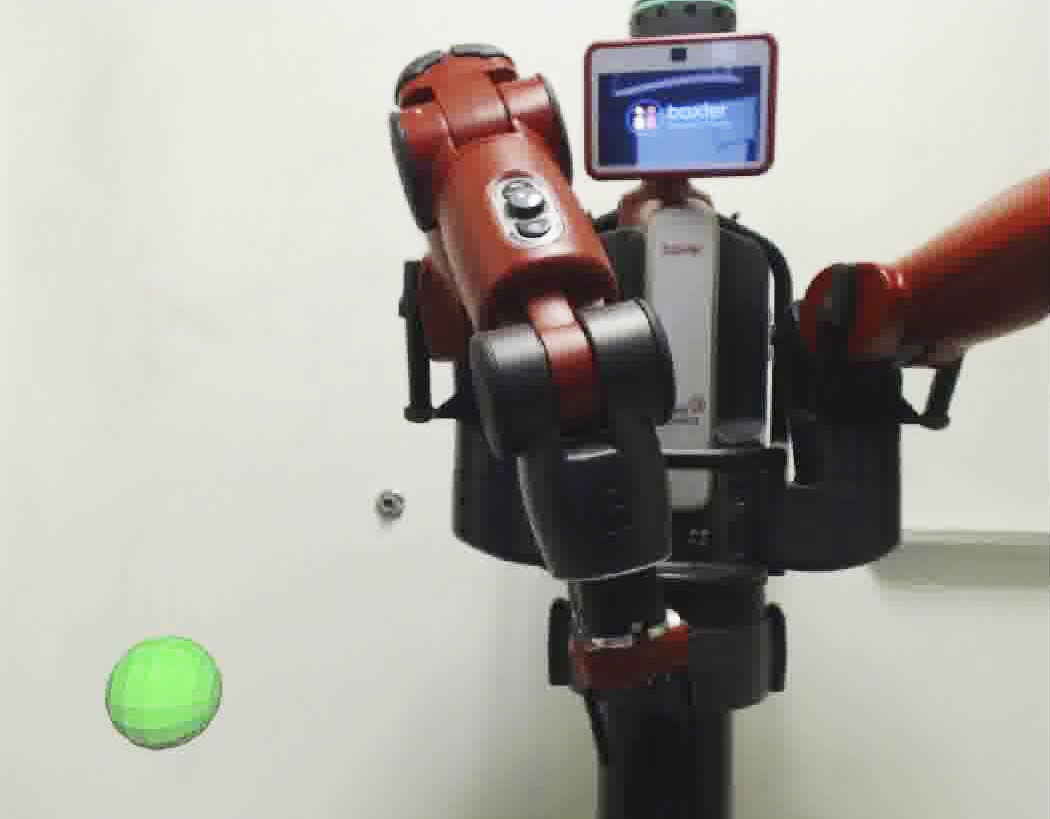} & 
    \hspace{-1.2em} \includegraphics[width=0.48\columnwidth]{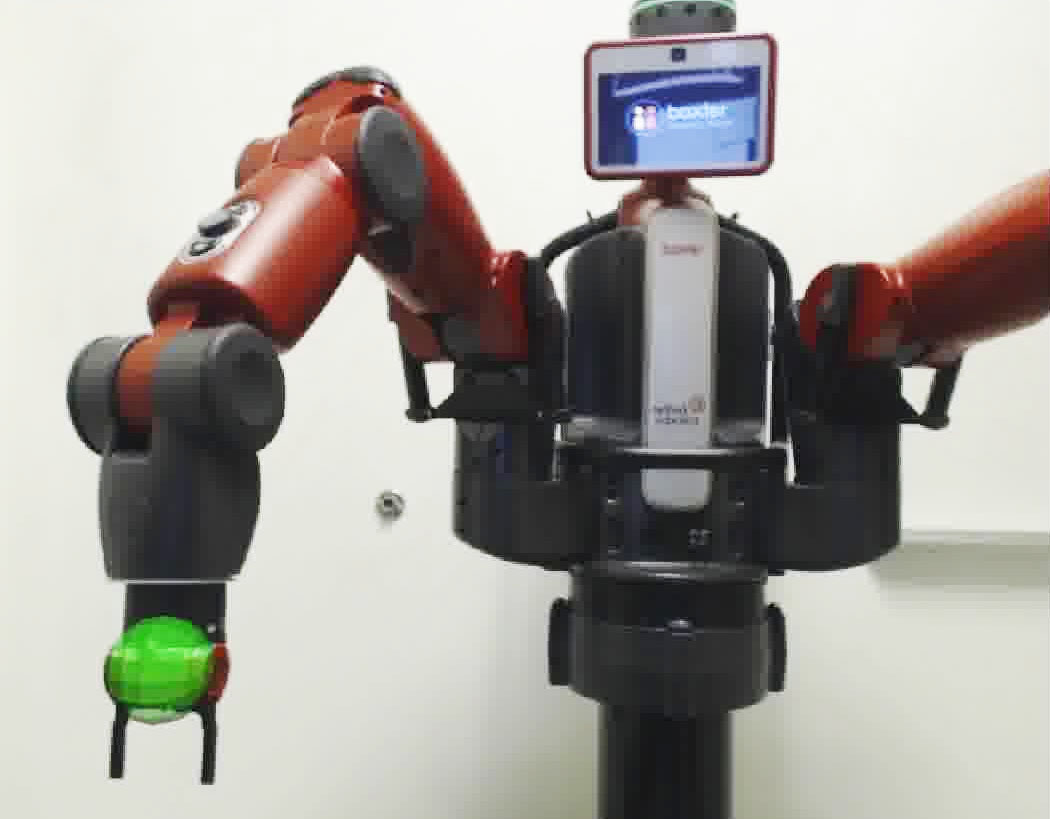} \\
    start position & end position
    \end{tabular}
    \caption{Real Baxter robot reaching a sequential goal position (green sphere), successfully overlapping the wrist with the goal, and achieving zero velocity.}
    \label{fig:baxter_greenball}
\end{figure}

Fig.~\ref{fig:baxter_time_plot} shows a 
plot of the step response as the robot moved from a start position to a goal position more than 30~cm away.  
Notice the relatively smooth behavior, and the moderate amount of overshoot and oscillation.
Similarly, Fig.~\ref{fig:baxter_greenball} shows one example of the robot moving between two sequential goal positions.

A compelling advantage of the proposed JAiLeR approach is that learning on the robot comes essentially for free.
That is, the same code that we use for training in simulation can be applied to the real robot.
This fine-tuning procedure can be used to reduce the errors even further.
Note that this is contrary to popular wisdom, which says that model-free reinforcement learning should not be applied to a real robot due to sample inefficiency.
In our case, however, because the policy has already been trained in simulation, fine-tuning on the real robot is safe and practical.
We fine-tuned the policy on the Baxter continuously for several hours by running PPO, sampling goal positions sequentially in the manner described above.
We used the same training code as in simulation, except that the penalty term for acceleration was decreased by a factor of two to encourage the robot
to reach the target.
Fine-tuning proceeded for up to 4000 network updates (iterations).
With this procedure, the error was reduced even further, yielding an average error of 0.7~cm.

In addition, an early version of our system was transferred to a real Kiova Jaco robot and used to control the robot to grab bottles on a table and pour them, using prespecified waypoints with orientation.
Another policy was transferred to the Kinova Gen3 robot, achieving 0.5~cm average accuracy in a relatively large workspace.\footnote{The workspace was 120$^\circ$ of an annular cylinder with height 0.5~m and inner/outer radii of 0.45 and 0.85~m, respectively.}

\section{Relationship to previous work}

Using a neural network to control a robot arm is not new. Several researchers have explored this idea for a small number (2--3) of joints since the early 1990s.
Martinetz \emph{et al.}~\cite{martinetz1990tnn:3d} learned visuomotor control (inverse kinematics) of a simulated 3-DoF robot arm with a pair of simulated cameras for sensing.  The network achieved a positioning error of 0.3\% of the workspace dimensions, which matches our results.
DeMers \emph{et al.}~\cite{demers1992nips:ik} learned inverse kinematics for a 3-DoF manipulator in 3D space using a combination of unsupervised and supervised learning; the approach only works for non-redundant manipulators.
Smagt \emph{et al}.~\cite{smagt1991icann:robcon} learned to control 2 joints of a 6-DoF robot arm to hover above an object using a camera-in-hand, with only tens of trials. %
Sanger \cite{sanger1994tra:gradual} proposed ``trajectory extension learning''---a curriculum learning strategy---to train a neural network to learn inverse dynamics to follow specific trajectory.  The technique was applied to a 2-DoF real arm and a 3-DoF simulated arm. 
Recent research has also explored the reaching problem for low degrees of freedom in simulation~\cite{brockman2016openai,schulman2017proximal}.

In the late 2000s, Peters and Schaal~\cite{peters2006rss:opctrl,peters2007icra:opctrl,peters2008ijrr:opctrl} learned joint space control to track Cartesian targets by learning piecewise linear regions. 
Their approach was able to learn a specific trajectory, but was not capable of reaching arbitrary goals within the workspace.
Similarly, other researchers have explored learning in the context of specific tasks~\cite{levine2015,chebotar2019icra:simopt}.

Recently, Mart{\'i}n-Mart{\'i}n \etal~\cite{martin2019variable} explored different action representations for manipulation. 
Their study shows the difficulty of learning in joint space, where their path-following policy is able to reach within 5~cm of a fixed sequence of four Cartesian points with an accuracy of about 70\%.
In an approach similar to ours but for a different task,  Hwangbo~\etal~\cite{hwangbo2019sr:agile} use deep RL in simulation to learn policies that transfer to the real world, for quadruped walking.

A closely related problem is that of learning the inverse kinematics of the robot, then using a joint controller to reach the inverse mapped joint configuration.
Several researchers have explored this idea~\cite{bocsi2011iros:invkin,Danieletal}.
The work of D'Souza~\etal~\cite{dsouza2001iros:invkin} is most closely related to ours, in that they approximate the nonlinear mapping with local piecewise linear models, and they learn incrementally.  
In our case, the deep neural network automatically learns such local models, and curriculum learning is used to learn incrementally.

Despite the promise of deep RL, most recent approaches in this area are limited to operating in end-effector space to execute learned manipulation tasks~\cite{sadeghi2018cvpr:sim2real,shao2020icra:scaffold,lee2020guided,lee2019icra:vistouch,Kalashnikov2018corl:qtopt}. 
Lower-level control of the robot joints is then achieved by traditional operational space controllers.
This choice of state space prevents such approaches from performing tasks that involve avoiding arm contact, such as reaching in clutter.

As discussed earlier, Lewis~\etal~\cite{lewis2019tr:drl} study a problem similar to ours.  They use a deep RL algorithm (DDPG) to learn to reach any goal position within a workspace.
Their work differs in several respects, namely, that their start position is fixed, and success is declared as soon as the end effector reaches within a radius of the goal, without requiring it to stop at the goal, similar to standard RL benchmarks~\cite{brockman2016openai}.
Even so, their average error is about an order of magnitude worse than ours (approximately 10~cm versus 1~cm).

More recently, Aumjaud~\etal~\cite{aumjaud2020arx:rlreach} study the ability of deep RL algorithms (including PPO) to solve the reaching problem.  With a fixed start and goal, they find that most algorithms are able to successfully achieve an error less than 5~cm in simulation, and about half the algorithms are successful in the real world.  Performance, however, drops precipitously when the goal position is randomized (even when the start remains fixed), with most algorithms (including PPO) achieving less than 10\% success at a threshold of 5~cm (the best algorithm achieves 75\% success).
Additionally, all algorithms struggle to achieve any measurable success either in simulation or the real world at a threshold of 1~cm.
In contrast with these results, we show that deep RL is able, via curriculum learning, to learn a general-purpose reaching controller (from a random start to a random goal) with high accuracy (less than 1~cm error) within a large workspace, both in simulation and in reality.

\section{Conclusion}

We presented JAiLeR, a deep RL-based approach to controlling a robot manipulator at the joint level that exhibits accurate, smooth, and stable motion.
Trained in simulation with curriculum learning, this simple approach is able to achieve reaching accuracy comparable to that of classical techniques (sub-centimeter) over a large workspace, in both simulation and reality.  
Advantages of this learned approach include automatic handling of redundancy, joint limits, and acceleration / deceleration profiles.
Moreover, fine-tuning directly on the real robot is feasible, thanks to transfer learning.
By augmenting the input to the network, 
the same approach can be used to reach while avoiding obstacles.
We hope these encouraging results inspire further work in this area.
Future research will connect the learned controller with real-time perception.

\section{Appendix:  Operational space control}

For more than three decades, the dominant approach to controlling redundant robots has been operational space control~\cite{khatib1987ra:osc}.
In this section we provide background material and notation related to this approach.

\subsection{Basics}

A robot manipulator with $n$ rotational joints is controlled by sending torques~$\bftau \in \R^n$ to the motors. These torques can be computed via the inverse dynamics control law~\cite{de2012theory} as
\begin{align*}
  \bftau &= \bfM(\bfq)\ddot{\bfq}_r + \bfC(\bfq,\dot{\bfq})\dot{\bfq} + \bfg(\bfq), \numberthis{eq:control_law}
\end{align*}
where $\bfq, \dot{\bfq}, \ddot{\bfq} \in \R^n$ are the joint angles, velocities, and accelerations, respectively; $\bfM(\bfq),\bfC(\bfq,\dot{\bfq}) \in \R^{n \times n}$ and~$\bfg(\bfq) \in \R^n$ are the manipulator's inertial matrix, the Coriolis forces, and gravitational force, respectively; and $\dot{\bfq}_r, \ddot{\bfq}_r \in \R^n$ are the reference velocity and acceleration, respectively. Despite the fact that \eqnref{eq:control_law} is an approximation that does not take into account the effects of joint friction or motor dynamics, this widely-used model has been found in practice to be sufficient for controlling manipulators. 

For many manipulation tasks, we are interested in reaching a goal configuration~$\bfx_d$ for the end-effector in Cartesian space (also known as \emph{task space}, or \emph{operational space}). This goal configuration can be a Cartesian position~($\bfx_d\in \mathbb{R}^3$), a full Cartesian pose~($\bfx_d\in \mathbb{SE}(3)$), or a position with orientation constraints. 
To solve this problem, the task space goal must be mapped to joint space.
When the number of degrees-of-freedom is greater than the dimensionality of the task space, an infinite number of joint configurations will satisfy any particular goal configuration.
This redundancy is often leveraged in manipulation tasks to satisfy other task requirements such as avoiding collisions.

The fundamental relationship between joint and task space velocities is ${\dot \bfx} = \bfJ(\bfq) {\dot \bfq}$, where~$\bfJ(\bfq)$ is the kinematic Jacobian of the manipulator.  Differentiating this equation yields the corresponding relationship between accelerations in the two spaces:
\begin{align*}
    \ddot{\bfx} &= \bfJ \ddot{\bfq} + \dot{\bfJ} \dot{\bfq}, \numberthis{eq:xjqjq}
\end{align*}
where the dependency of $\bfJ$ on $\bfq$ has been omitted for brevity.

Given a goal position, velocity, and/or acceleration in task space, we can use the equations above to compute the corresponding quantities in joint space.
Then, given a desired 
position~$\bfq_d$, 
velocity~$\dot{\bfq}_d$, and
acceleration~$\ddot{\bfq}_d$ in joint space, the reference acceleration can be computed as
\begin{align*}
  \ddot{\bfq}_r &= k_p(\bfq_d - \bfq) + k_d(\dot{\bfq}_d-\dot{\bfq}) + \ddot{\bfq}_d, \numberthis{eq:reference_acceleration}
\end{align*}	
where~$k_p, k_d \in \R$ are the proportional and derivative gains, respectively.\footnote{In practice, these gains are often diagonal matrices, to allow different scales for different joints.} 
The gain~$k_p$, also known as the ``stiffness gain", penalizes any deviation from the desired joint position~$\bfq_d$; whereas $k_d$, often referred to as the ``damping gain", dampens any sudden movement of the robot from its desired velocity~$\dot{\bfq}_d$. 

Since the reference acceleration is premultiplied by the robot's inertial matrix in \eqnref{eq:control_law}, any error in the robot dynamics model will affect the robot's ability to reach the target. 
To alleviate this issue in real robotic systems, the control law obtained by combining \eqnsref{eq:control_law}{eq:reference_acceleration} is often modified to remove the premultiplication by $\bfM$ from the proportional and derivative terms:
\begin{align*}
\bftau &= \bfM\ddot{\bfq}_d + \bfC\dot{\bfq}_d + \bfg + k_p(\bfq_d - \bfq) + k_d (\dot{\bfq}_d - \dot{\bfq}), \numberthis{eq:inv_dyn_law}
\end{align*}
where for notational brevity the dependency of $\bfM$, $\bfC$, and $\bfg$ on the joint angles and velocities has been omitted.

\subsection{Controllers}
\label{subsec:controllers}

Nakanishi~\etal~\cite{schaalosc} analyze the eight most common ways, in the context of operational space control (OSC), to map task space goals to joint torques to move the robot's end-effector to follow a particular Cartesian trajectory. 
The two best performing methods from their work are as follows.

\textbf{1) Velocity-based controller (OSC-V).}\footnote{This controller is \#2 in \cite{schaalosc}, described in \S3.1.2 of that paper.} Given a target position~$\bfx_d$ and velocity~${\dot \bfx}_d$ in Cartesian task space, this controller first computes a reference velocity $\dot{\bfx}_r$ using a proportional-derivative (PD) controller:
\begin{align*}
\dot{\bfx}_r &= \kappa_p (\bfx_d - \bfx) + {\dot \bfx}_d, \numberthis{eq:osc_vel}
\end{align*}
where~$\kappa_p$ is the stiffness gain in task space.

The reference joint velocity~$\dot{\bfq}_r$ is then computed as
\begin{align*}
\dot{\bfq}_r &= {\bfJ^+} {\dot \bfx}_r - \alpha (\bfI - \bfJ^+ \bfJ) \bff, \numberthis{eq:oscvel_refvel}
\end{align*}
where~$\bfJ^+(\bfq)$ is the Moore-Penrose pseudoinverse of the Jacobian; and $\bff \in \R^n$ is the null space force, which allows for additional task goals such as avoiding collisions and avoiding singularity configurations of the robot. 

To follow this reference joint velocity~$\dot{\bfq}_r$, along with the reference joint acceleration~$\ddot{\bfq}_r$ (obtained by differentiating the reference velocity), the controller computes
\begin{align*}
\bftau &= \bfM {\ddot \bfq}_r + \bfC {\dot \bfq}_r + \bfg + k_d (\dot{\bfq}_r - \dot{\bfq}), \numberthis{eq:osc_vel_law}
\end{align*}
which is similar to \eqnref{eq:inv_dyn_law} except that the velocity is not integrated to obtain a reference position (thus $k_p=0$).

\textbf{2) Simplified acceleration-based controller (OSC-A).}\footnote{This controller is \#5 in \cite{schaalosc}, described in \S3.2.3 of that paper.}
Given a target position~$\bfx_d$, velocity~${\dot \bfx}_d$, and acceleration~${\ddot \bfx}_d$, this controller first computes the reference acceleration in task space as
\begin{align*}
  \ddot{\bfx}_r &= \kappa_p (\bfx_d - \bfx) + \kappa_d (\dot{\bfx}_d - \dot{\bfx}) + \ddot{\bfx}_d. \numberthis{eq:osc_acc}
\end{align*}
The reference joint acceleration is then obtained by rearranging \eqnref{eq:xjqjq}:
\begin{align*}
    \ddot{\bfq}_r &= \bfJ^+ (\ddot{\bfx}_r -  \dot{\bfJ}\dot{\bfq}), \numberthis{eq:refjointaccel}
\end{align*}
assuming that the actual joint velocity well approximates the reference joint velocity, $\dot{\bfq} \approx \dot{\bfq}_r$.

To follow this reference acceleration, the controller computes
\begin{align*}
    \bftau &= \bfM \ddot{\bfq}_r + \bfC {\dot \bfq} + \bfg - (\bfI - \bfJ^+ \bfJ) (k_d \dot{\bfq} + \alpha \bff), \numberthis{eq:osc_acc_law}
    \end{align*}
which is similar to \eqnref{eq:osc_vel_law} except that the null space is not premultiplied by $\bfM$, and the null space includes an extra term to dampen the torques based on the velocities.

\bibliographystyle{IEEEtran}
\bibliography{root}

\end{document}